\documentclass[conference]{IEEEtran}
\IEEEoverridecommandlockouts
\usepackage{multirow} 
\usepackage{threeparttable} 

\usepackage{times}
\usepackage{soul}
\usepackage{url}
\usepackage[hidelinks]{hyperref}
\usepackage[utf8]{inputenc}
\usepackage[small]{caption}
\usepackage{graphicx}
\usepackage{amsmath}
\usepackage{amsthm}
\usepackage{booktabs}
\usepackage{algorithm}
\usepackage{algorithmic}
\usepackage[switch]{lineno}
\usepackage{cite}
\usepackage{amsmath,amssymb,amsfonts}
\usepackage{algorithmic}
\usepackage{graphicx}
\usepackage{textcomp}
\usepackage{xcolor}
  {}

\def\BibTeX{{\rm B\kern-.05em{\sc i\kern-.025em b}\kern-.08em
    T\kern-.1667em\lower.7ex\hbox{E}\kern-.125emX}}
\begin{document}

\title{Enhancing Large-scale UAV Route Planing with Global and Local Features via Reinforcement Graph Fusion
\thanks{This work was supported by the National Natural Science Foundation of China under Grant 62276222.\\
\hspace*{\parindent}*Corresponding author: Kai Ye.
}
}

\author{\IEEEauthorblockN{1\textsuperscript{st} Tao Zhou}
\IEEEauthorblockA{\textit{ School of Informatics} \\
\textit{Xiamen University}\\
Xiamen, China \\
taozhou@stu.xmu.edu.cn}
\and
\IEEEauthorblockN{2\textsuperscript{nd} Kai Ye \IEEEauthorrefmark{1}}
\IEEEauthorblockA{\textit{ School of Informatics} \\
\textit{Xiamen University}\\
Xiamen, China \\
yekai@stu.xmu.edu.cn}
\and
\IEEEauthorblockN{3\textsuperscript{rd} Zeyu Shi}
\IEEEauthorblockA{\textit{Marine Design and Research Institute of China} \\
Shanghai, China \\
shizeyu@maric.com.cn}
\and
\IEEEauthorblockN{4\textsuperscript{th} Jiajing Lin}
\IEEEauthorblockA{\textit{ School of Informatics} \\
\textit{Xiamen University}\\
Xiamen, China \\
31520231154298@stu.xmu.edu.cn}
\and
\IEEEauthorblockN{5\textsuperscript{th} Dejun Xu}
\IEEEauthorblockA{\textit{ School of Informatics} \\
\textit{Xiamen University}\\
Xiamen, China \\
xudejun@stu.xmu.edu.cn }

\and
\IEEEauthorblockN{6\textsuperscript{th} Min Jiang}
\IEEEauthorblockA{\textit{ School of Informatics} \\
\textit{Xiamen University}\\
Xiamen, China \\
minjian@xmu.edu.cn}}

\maketitle

\begin{abstract}
Numerous remarkable advancements have been made in accuracy, speed, and parallelism for solving the Unmanned Aerial Vehicle Route Planing (UAVRP). 
However, existing UAVRP solvers face challenges when attempting to scale effectively and efficiently for larger instances. 
In this paper, we present a generalization framework that enables current UAVRP solvers to robustly extend their capabilities to larger instances, accommodating up to 10,000 points, using widely recognized test sets. The UAVRP under a large number of patrol points is a typical large-scale TSP problem.Our proposed framework comprises three distinct steps. 
Firstly, we employ Delaunay triangulation to extract subgraphs from large instances while preserving global features. 
Secondly, we utilize an embedded TSP solver to obtain sub-results, followed by graph fusion. Finally, we implement a decoding strategy customizable to the user's requirements, resulting in high-quality solutions, complemented by a warming-up process for the heatmap. 
To demonstrate the flexibility of our approach, we integrate two representative TSP solvers into our framework and conduct a comprehensive comparative analysis against existing algorithms using large TSP benchmark datasets. 
The results unequivocally demonstrate that our framework efficiently scales existing TSP solvers to handle large instances and consistently outperforms state-of-the-art (SOTA) methods. 
Furthermore, since our proposed framework does not necessitate additional training or fine-tuning, we believe that its generality can significantly advance research on end-to-end UAVRP solvers, enabling the application of a broader range of methods to real-world scenarios.
\end{abstract}

\begin{IEEEkeywords}
UAV, Route Planning, TSP, Large-scale, Reinforcement Graph Fusion
\end{IEEEkeywords}

\section{Introduction}
With the development of  Unmanned Aerial Vehicle (UAV), using UAV for patrolling has become an emerging and important security measure. Planning a reasonable patrol route is key to improving the efficiency and effectiveness of the patrol. Our approach is to design a path planning algorithm for UAV to enhance patrol efficiency. 
The Unmanned Aerial Vehicle Route Planing
(UAVRP) is a optimization problem wherein a UAV must visit a specific set of sites exactly once and return to the starting point, aiming to minimize the overall travel distance.
The primary hurdle in UAVRP's resolution lies in the extensive search space that emerges when dealing with a substantial number of sites, denoted as $n$. The UAVRP under a large number of
patrol sites is a typical  Large-Scale Travelling Salesman Problem (LSTSP), we define a TSP instance that requires visiting more than 500 sites as an LSTSP in this paper.
Despite its inherent theoretical complexity, the LSTSP finds numerous practical applications across various domains such as drone delivery, transport, genome sequencing, and circuit board design \cite{agatz2018optimization,borkakoty2021tsp,wang2024spatial,xu2024efficient,zhang2022fast,jiang2012fuzzy}.

Over the years, the operations research community has proposed various methods to tackle the UAVRP.
Among these approaches, Concorde \cite{APPLEGATE200911} stands out as a relatively powerful and extensively utilized method in real-world scenarios.
However, when confronted with UAVRP problems involving tens of thousands of dimensions, Concorde falls short in producing accurate results within a reasonable timeframe (within 8 CPU hours). 
To address this limitation, numerous heuristic algorithms have been developed, including LKH3 \cite{36b9628e7a874d208624584d8a470985} and OR-Tools. 
Nevertheless, when dealing with the LSTSP, these algorithms suffer from two key drawbacks.
Firstly, they are still time-consuming. 
Secondly, their iterative search procedures yield unstable results and necessitate the manual development of heuristic rules.

To tackle the challenges of time-consuming and unstable solutions, a series of learning-based methods have been proposed \cite{DBLP:conf/acml/CostaRZA20,kool2018attention,hong2024boosting}.
Such methods are trained on huge volumes of data to obtain features of instances from different paradigms. 
These methods are trained on extensive datasets to extract features of instances from various paradigms, enabling them to handle previously unseen instances with fast inference speed and search stability, resulting in excellent performance on the LSTSP.\cite{jiang2020fast}

Learning-based TSP approaches fall into one-step and two-step categories based on solution steps.

One-step methods, including construction-based and iterative, find solutions directly without intermediates. They perform well on small TSPs but have difficulty scaling to LSTSPs efficiently and stably \cite{jiang2020individual,tan2021evolutionary}.

Two-step methods generate intermediates like heatmaps indicating edge probabilities in the optimal solution and then search these to find the final solution. \cite{DBLP:conf/aaai/FuQZ21}, for example, uses a supervised model for sub-heatmaps and Monte Carlo tree search \cite{wang2024multiview}. While this generalizes to 10,000-dimensional LSTSPs and maintains competitive accuracy within a reasonable time, the divide-and-conquer approach can lose global features during sub-graph extraction, leaving room for improvement.

There's a notable gap in generalizing from TSP to LSTSP, with algorithms facing high computational costs and accuracy loss \cite{hong2024efficiently}. This gap means many promising TSP solvers are underutilized in realistic, large-scale scenarios.

To overcome these challenges, we propose a framework for TSP solvers generalization without training, efficiently extending existing TSP solvers to handle LSTSP instances and achieving results comparable to state-of-the-art methods. The proposed Delaunay Triangulation-based TSP-solver Generalization Framework (DTTGF) operates as depicted in Figure \ref{Pipline} for instances of the LSTSP.
For an LSTSP instance, our framework first triangulates it, enabling subgraph extraction to preserve global features.
The extracted sub-graphs are then solved by the embedded TSP solver to obtain sub-products, such as sub-solutions or sub-heatmaps.
These sub-products are merged to generate the heatmap of the original instance. 
To improve sub-graph fusion, we propose a pseudo-reinforcement learning based warm-up mechanism that removes misleading edges while correcting the heatmap.
Finally, the generated heatmap is searched using widely adopted search methods to obtain the final solution. 
Our divide-and-conquer based framework ensures adaptability and reliable generalization for all existing TSP solvers.

The contributions are summarized as follows:
\begin{itemize}
\item We introduce a novel approach based on triangulation for dividing subgraphs. 
This method ensures that global features within the subgraphs are preserved while maintaining the overall structure intact.
\item We propose an innovative warm-up strategy based on pseudo-reinforcement learning. This strategy enhances search outcomes while incurring an acceptable trade-off in terms of time efficiency.
\item We develop a unified framework for generating sub-graphs. This framework enables the integration of diverse solvers rooted in reinforcement learning and supervised learning methodologies. The resultant solvers can seamlessly adapt to instances of varying sizes. 
\end{itemize}
The work addresses the need for scalable solutions for both current and potential future TSP solvers. This adaptability ensures that existing small-scale accurate TSP solvers can effectively extend their applicability to real-world scenarios. You can find our code and appendix on GitHub above:https://github.com/Calomiya/DTTGF.git.

\section{Related works}
The TSP problem, being a long-standing NP-hard challenge, has garnered a substantial body of related research.  
Given the primary focus of our paper on designing a framework for learning-based approaches, we delve into the intricate details of these approaches in the subsequent sections, rather than attempting to encompass all domains.
Moreover, the landscape boasts several exceptional algorithms in diverse fields, including evolutionary computation. 
Those interested in non-learning methods can delve deeper into these alternatives in \cite{osaba2020traveling,jiang2017integration,jiang2017transfer,wang2024evolutionary}.
\subsection{Learning-based Methods}
We classify learning-based strategies into one-stage and two-stage methods, contingent on the progression stages required to achieve a solution and the necessity for intermediate product generation.
\subsubsection{One-stage Methods}
Point Networks \cite{vinyals2015pointer,jiang2020knee} pioneered an end-to-end TSP approach using neural attention to handle variable output sizes, selecting inputs as outputs.
\cite{kool2018attention} improved this with an attention layer, outperforming pointer networks, and used REINFORCE for model training with a greedy rollout baseline.
POMO \cite{kwon2020pomo} explored REINFORCE for guiding TSP solutions towards multiple optima \cite{rambabu2019mixture}.
\cite{pan2023h-tsp} proposed a direct TSP solution approach, avoiding costly search processes, offering a fast solution but challenging to generalize to Large-Scale TSP without extensive fine-tuning or retraining.

\subsubsection{Two-stage Methods}
\cite{DBLP:conf/aaai/FuQZ21} introduces a two-stage learning-based TSP solver: generating an intermediate heatmap followed by solution search. Using divide-and-conquer, it breaks down LSTSP into smaller instances, solves sub-heatmaps, and merges results, then applies MCTS for performance comparable to LKH3 on 10,000-point instances. The approach is adaptable and scalable despite the additional search phase.

DIMES \cite{qiu2022dimes} follows this model, offering two-stage processing with heatmap generation and MCTS for improved accuracy. It also creates a continuous space for parameterizing solution distributions, which stabilizes REINFORCE training and enables parallel sampling for fine-tuning.

It is important to underscore that the aforementioned  DIMES \cite{qiu2022dimes}, H-TSP \cite{pan2023h-tsp} and Att-GCN \cite{DBLP:conf/aaai/FuQZ21} demonstrate the capacity to generalize to LSTSP.  
However, the scope of generalization for these three solutions remains confined to their respective models and does not extend to other pre-existing TSP solvers.

\begin{figure*}[t]
\centering
\includegraphics[width=0.9\textwidth]{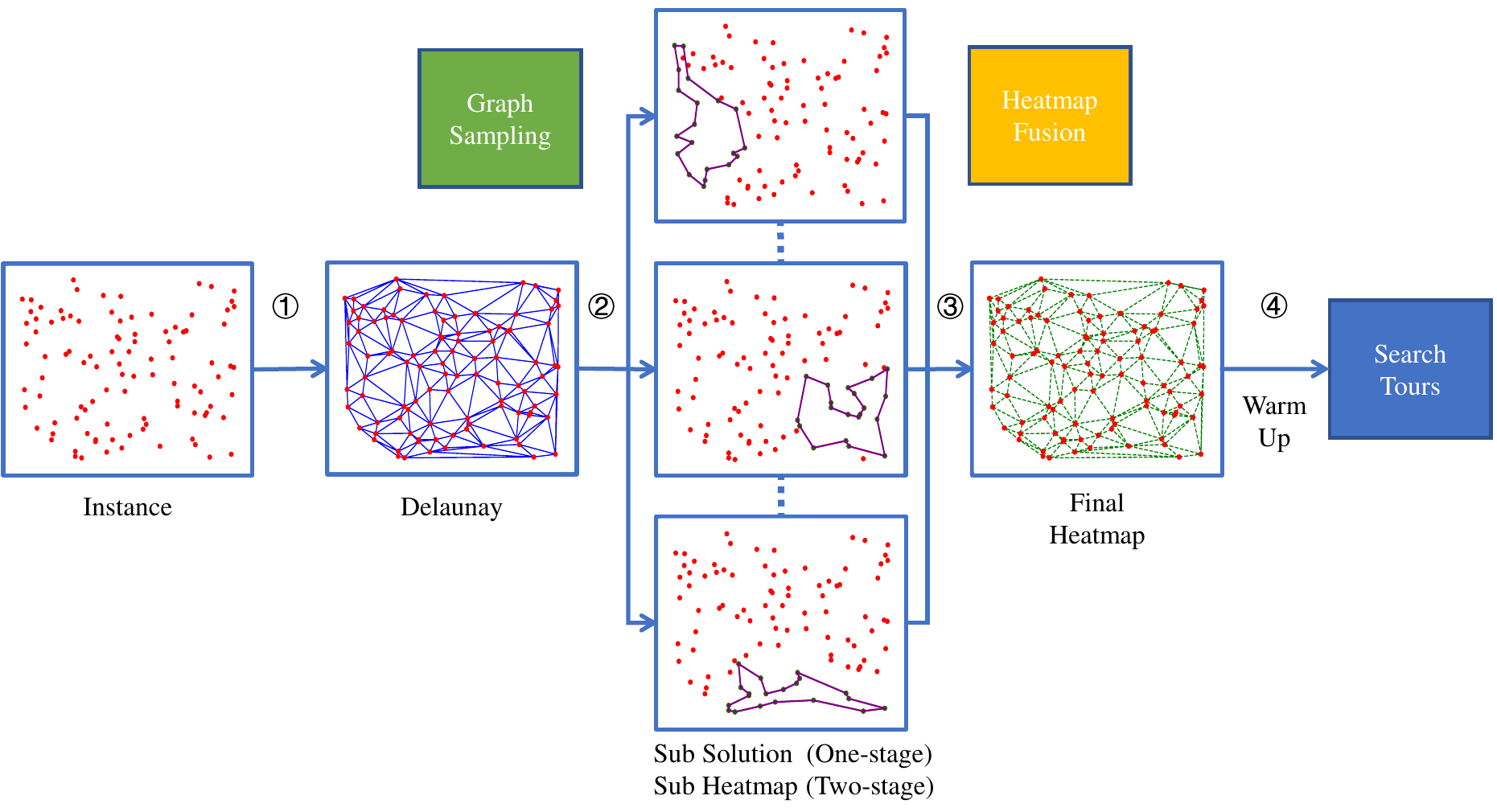} 
\caption{Pipeline of the proposed approach.
Step 1: perform Delaunay Triangulation(DT) on the current instance. 
Step 2: sampling and solving the sub-graph on DT results.
Step 3: merging sub-solutions (one-stage solver) or sub-heatmaps (two-stage solver) to obtain a global heat map. 
Step 4: warming up heat map and search for the global optimal solutions.}
\label{Pipline}
\end{figure*}

\section{Methods}
\subsection{Preliminaries}
\subsubsection{Problem Definition}
Our proposed framework concentrates on the Two-Dimensional Euclidean Traveling Salesman Problem (2D Euclidean TSP). 
This problem can be represented as an undirected graph $G(V,E)$, where $V$ ($|V|=n$) signifies the set of vertices, referred to as cities in the context of TSP, and $E$ represents the set of edges.
For consistency with the majority of learning-based methodologies, we presume that all vertices in $V$ are uniformly distributed within a unit square of side length 1. 
This translates to the coordinates $(x_i, y_i)$ of each vertex adhering to the condition $x_i\in[0,1],y_i\in[0,1]$.
The Euclidean distance between cities is denoted by $d_{ij}$, and an element $p_{ij}$ within the heatmap $P$ for graph $G$ signifies the probability of edge $(i,j)$ being present in the optimal solution.

\subsubsection{Delaunay Triangulated Graph}
Delaunay Triangulation(DT) is the fundamental method of study in algebraic topology. 
In the case of a surface, we divide the surface into pieces that satisfy the following conditions:
\begin{itemize}
\item Each piece is a curved triangle; 
\item Any two such curved triangles on the surface either do not intersect or intersect on exactly one common side (not two or more sides at the same time).
\end{itemize}

The definition of DT requires it to have the following characteristics:
\begin{itemize}
\item The fact that the outer circle of every Delaunay triangle contains no other point in its area is called the empty outer circle property of Delaunay triangles, and this property has been used as a criterion for creating Delaunay triangles. 
\item Another property is the maximum minimum angle property: the minimum angle of the six interior angles of a convex quadrilateral formed by the diagonals of any two adjacent triangles does not increase when they are exchanged.
\end{itemize}

Several heuristic TSP solvers \cite{9007037,Letchford_Pearson_2008,alkema_et_al:LIPIcs.ESA.2022.5} have empirically demonstrated that the optimal solution of a TSP is highly likely to reside along the edges delimited by the DT. 
We incorporate this attribute into our learning-based methodology to uphold global features throughout the process of sub-graph extraction.

\subsection{The Proposed Framework}
The DTTGF process encompasses 7 primary steps as depicted in Figure \ref{Pipline}. 
Initially, the Delaunay Triangulation (DT) is executed on the instances, followed by the extraction of sub-graphs based on the DT outcomes. 
The derived sub-graphs are then addressed by the integrated TSP solver to yield sub-results, encompassing sub-heatmaps or sub-solutions. 
Subsequently, these sub-results amalgamate to construct a global heatmap, which is subsequently processed warming-up strategy to generate the ultimate heatmap. 
The final heatmap undergoes a search process utilizing the chosen search technique, culminating in the production of the definitive solution.

In this paper, as a framework-oriented study, we integrate the most distinct S+2-OPT (Sampling Decoder+2-OPT) \cite{kool2018attention,Lin_Kernighan_1973} and MCTS (Monte Carlo Tree Search) \cite{DBLP:conf/aaai/FuQZ21} approaches into the proposed DTTGF during the experimental phase to showcase the heatmap's performance.

 \begin{algorithm}[htb]
  \caption{Procedure of DTTGF}
  \label{alg:over}
  \small
  \begin{algorithmic}[1]
    \REQUIRE
      TSP Instance, $V=\{v_1,v_2,...,v_N\}$;
      Selected TSP-solver, $S$
    \ENSURE
      Solution route, $\tau=\{\tau_1,\tau_1,...,\tau_N\}$;
    \STATE $V_D\leftarrow$$Delaunay Triangulation(V)$;
    \STATE $V_{sub}\leftarrow$$Graph Sampling(V_D)$;
    \STATE $\{P_{sub}^1,P_{sub}^2,...,P_{sub}^M\}\leftarrow$$SubGraphSolving(V_{sub},S)$;
    \STATE $P\leftarrow$$GraphMerging(\{P_{sub}^1,P_{sub}^2,...,P_{sub}^M\})$;
    \STATE $P_W\leftarrow$$WarmingUp(P)$;
    \STATE $\tau\leftarrow$$TourSearching(P_W,V)$;
    \RETURN $\tau$;
  \end{algorithmic}
\end{algorithm}

\subsection{Graph Sampling}
For LSTSP, the divide-and-conquer strategy is common. \cite{DBLP:conf/aaai/FuQZ21} uses k-nearest neighbors for sub-graph decomposition to reduce complexity, but this risks losing global features. This is crucial in TSP, as sub-graph solutions must align with the overall solution. Our DT-centric approach ensures effective sub-graph extraction while preserving global properties.

Instances are processed with DT before sub-graph extraction. The DT result is represented in an adjacency matrix $D$, where $D_{ij} = 1$ for DT edges and $D_{ij} = 0$ otherwise. To cover the entire graph evenly, DTTGF uses a matrix $O$ of size $n$, incrementing $O_i$ each time node $i$ is selected for a sub-graph. The framework starts with the node $i$ having the lowest $O_i$ and explores adjacent nodes using a depth-first search, sorting them by distance $d_{ik}$.

Starting with the first node $k_1$ from the sorted list, it is included in the sub-graph, and $O_{k1}$ is incremented. The process continues, adding connected nodes from $D$ and selecting the closest to extend the sub-graph, until the node count is met. Iteration stops when sub-graphs meet the criteria.

\subsection{Sub-graph Solving}
The difference between one-stage and two-stage solvers is the generation of an intermediate heatmap.
Learning-based approaches share fast search speeds through parallelism or strong forward processing, handling many small instances quickly.
However, they struggle with varying instance sizes in real-world scenarios, where parallelism is less effective.
Despite this, it forms the basis for our framework, leveraging fixed sub-graph sizes and fast search times to manage many sub-graphs efficiently, meeting real-world time needs.
This paper includes representative methods in DTTGF for both stages.
The one-stage solver provides sub-graph solutions, and the two-stage solver generates a heatmap, with the following section detailing how both are integrated into a global heatmap.

\subsection{Graph Merging}
\subsubsection{One-stage Solver}
For a one-stage solver, each sub-graph outputs a tour $T$, and the probability of each edge $(i,j)$ appearing in the optimal solution is as follows
\begin{equation}
P_{ij}=\frac{\sum_{l=1}^{I} T_l(i,j)}{S_{ij}}
\label{one-stage-p}
\end{equation}
where $I$ represents the count of sub-graphs, while $T_{l}(i,j)$ indicates the presence of edge $(i,j)$ within the solutions $T_l$ of sub-graph $l$.
Meanwhile, $S_{ij}$ signifies the aggregate instances of edge $(i,j)$ being chosen across all sub-graphs. As such, $P_{ij}$ denotes the actual frequency of edge $(i,j)$ within the solutions of all sub-graphs, calculated by dividing the occurrences of its selection in a sub-graph.

\subsubsection{Two-stage Solver}
The two-stage solver can also conduct dual search stages to obtain the final solution for the sub-graph and use equation (\ref{one-stage-p}) to compute the heatmap of the sub-graph, similar to the one-stage solver. 
However, the two-stage approach inherently benefits from the divide-and-conquer concept. 
This approach involves generating intermediate heatmaps itself, and if these intermediate heatmaps can be directly employed, the efficiency of producing global sub-graphs can be enhanced, mitigating the computational expense of the second search stage.

The corresponding heatmap for the two-stage sub-graphs is formulated as follows:
\begin{equation}
P_{ij}=\frac{\sum_{l=1}^{I} P_l(i,j)}{S_{ij}}
\label{two-stage-p}
\end{equation}
where $P_l(i,j)$ denotes the value of edge $(i,j)$ within sub-graph $l$ on the sub-heatmap.

The noteworthy distinction from the one-stage solver lies in the fact that $T_l$ exclusively contains information about the sub-solutions, while $P_l$ provided by the two-stage solver encompasses the probabilities of all sub-graph edges appearing in the optimal solution.
The proposed framework adeptly maps the outcomes of TSP solvers for sub-graphs onto the global heatmap for both scenarios.

It is crucial to note that techniques centered around sub-graph partitioning and fusion can potentially lead to the loss of global characteristics.  
Specifically, edges with elevated $P$ values might perform well within individual sub-graphs, or they could be selected only a relatively limited number of times and incidentally emerge in the optimal solution of those sub-graphs.
However, the latter case involves edges that are unlikely to be superior within the global graph. 
To address this, the paper introduces Delaunay Triangulation (DT) as a filter for the amalgamated heatmap.
As mentioned earlier, prior research has illustrated a pronounced correlation between the optimal TSP solution and the DT delineation outcome, indicating that edges in the optimal solution tend to be encompassed by the DT result \cite{gong2024cross}.
DTTGF capitalizes on this insight, and for the generated heatmap $P$, the $P$ values of edges $(i,j)$ not present in the DT result are set to 0. This adjustment is applied as follows:
\begin{equation}
P_{ij}=0,\enspace \forall(i,j)\notin \rm DT
\label{DT_P}
\end{equation}
After applying the filter, the proposed framework acquires the fused heatmap corresponding to the original instance.

\subsection{The Warming-up Strategy}
While DT as prior knowledge can eliminate certain non-potential edges, two issues still remain. 
Firstly, some potentially valuable edges that are not part of the DT result could be removed during the filtering.
Secondly, even the edges within the DT result might appear promising on a local scale but mislead on a global level.

Particularly, when integrating a one-stage solver, it characterizes only the edges of the optimal solution for each subgraph. 
This results in a sparse heatmap after fusion, where each edge's information is crucial.
In contrast, a two-stage solver provides a subgraph heatmap for each subgraph, containing information about the majority of edges in the subgraph. 
The fused heatmap it produces is relatively dense and can partially mitigate this issue.

To address these challenges, we introduce a warm-up strategy based on pseudo-reinforcement learning, primarily tailored to one-stage solvers.
During subgraph fusion, the heatmap is pruned to retain only edges with a potential to appear in the optimal solution.\cite{zeng2024ask,wang2024generating}
Two types of edges require special attention: those with longer distances $d_{ij}$ and those with higher values of $P_{ij}$.
In TSP, the final performance often hinges on the selection of longer edges in the optimal solution, and edges with higher $P_{ij}$ values have a greater likelihood of being selected. 
Thus, we define fitness values as follows:
\begin{equation}
A_{ij}=P_{ij}\times d_{ij}
\label{fitness}
\end{equation}
The warm-up process involves iterations, following this sequence.
The initial solution $T_{ori}$ is derived by rapidly solving the heatmap using a sampling decoder\cite{kool2018attention} followed by 2-OPT improvements (referred to as S+2-opt). 
Note the current solution $T_{ori}$ as the baseline, i.e. $T_b$.
For each iteration, we select the edge $(i, j)$ with the largest current $A_{ij}$, modify its $P_{ij}$ to zero, and apply S+2-opt to obtain the improved solution $T_{del}$. 
If $T_{del}$ outperforms the baseline $T_b$ (obtained from $T_{ori}$), $P$ is updated to $P_{del}$ after deleting the edge, and back-propagation is performed.
Let the back-propagation formula be

\begin{equation}
\begin{aligned}
P_{ij}&=P_{ij} + \alpha \times \beta(e^{\frac{D(T_b)-D(T_{del})}{D(T_d)}}-1) \\ 
\alpha&=0,\enspace \textrm{if}\enspace(i,j)\in (T_{b}\cap T_{del})\\
\alpha&=1,\enspace \textrm{if}\enspace(i,j)\in T_{b}\\
\alpha&=-1,\enspace \textrm{if}\enspace(i,j)\in T_{del}
\end{aligned}
\label{LOSS}
\end{equation}

where $D(T_b)$ denotes the length of the tour consisting of solving $T_b$, $D(T_{del})$ denotes the length of the tour consisting of solving $T_{del}$, $\beta$ is the learning rate, and the value of $\alpha$ depends on whether the edge $(i, j)$ belongs to $T_b$ or $T_{del}$ or not.
After back-propagation, if $T_{del}$ is better than $T_b$, $T_b$ is updated to $T_{del}$. Once the iteration requirements are met, the process concludes, and the final heatmap is output.

The proposed warm-up module selectively removes non-potential edges from focus in the heatmap.
Moreover, potential edges not in the DT results but filtered during fusion can enhance their $P$-value via back-propagation, increasing their chances of selection in the subsequent search process.

\section{Experiments}
Existing TSPsolver for UAVRP under a large number of sites, which is modeled as LSTSP.
To validate the efficacy of DTTGF, we conducted evaluations using three datasets generated by \cite{Fu_Qiu_Zha_2022}.
These datasets comprise 128 instances of 500 points, 128 instances of 1000 points, and 16 instances of 10000 points. 
This dataset selection ensures comparability with many current works on learning-based TSP solvers, which commonly use this dataset for training and validation.
All experimental results were acquired using machines equipped with one RTX 1080 and an Intel(R) Xeon(R) Gold 5118 CPU @ 2.30GHz (featuring 8 cores), and the results align with those of \cite{qiu2022dimes}.


\begin{table*}[htbp]
  \centering
  \caption{Results of DTTGF w.r.t. existing baselines, tested on TSP-500, TSP-1000 and TSP-10000.}
    \scalebox{0.65}{
    \begin{threeparttable} 
    \begin{tabular}{ll|ccc|ccc|ccc}
    \toprule
    \multirow{2}[2]{*}{Method} & \multicolumn{1}{c|}{\multirow{2}[2]{*}{Type}} & \multicolumn{3}{c|}{TSP-500} & \multicolumn{3}{c|}{TSP-1000} & \multicolumn{3}{c}{TSP-10000} \\
          &       & Length↓ & Drop↓ & Time↓ & Length↓ & Drop↓ & Time↓ & Length↓ & Drop↓ & Time↓ \\
    \midrule
    \multicolumn{1}{l}{Concorde} & OR(exact) & 16.55* & ---   & 37.66m & 23.12* & ---   & 6.65h & N/A   & N/A   & N/A \\
    \multicolumn{1}{l}{Gurobi} & OR(exact) & 16.55 & 0.00\% & 43.63h & N/A   & N/A   & N/A   & N/A   & N/A   & N/A \\
    \multicolumn{1}{l}{LKH-3} & OR    & 16.55 & 0.00\% & 46.28m & 23.12 & 0.00\% & 2.57h & 71.77* & ---   & 8.8h \\
    \multicolumn{1}{l}{Farthest Insertion} & OR    & 18.30 & 0.00\% & 0s    & 25.72 & 11.25\% & 0s    & 80.59 & 12.29\% & 6s \\
    \midrule
    \multicolumn{1}{l}{EAN(CPAIOR2018)} & RL+S  & 28.63 & 73.03\% & 20.18m & 50.30 & 117.59\% & 37.07m & N/A   & N/A   & N/A \\
    \multicolumn{1}{l}{EAN} & RL+S+2-OPT & 23.75 & 43.57\% & 57.76m & 47.73 & 106.46\% & 5.39h & N/A   & N/A   & N/A \\
    \multicolumn{1}{l}{AM(ICLR2019)} & RL+S  & 22.64 & 36.84\% & 15.64m & 42.80 & 85.15\% & 63.97m & 431.58 & 501.27\% & 12.63m \\
    \multicolumn{1}{l}{AM} & RL+G  & 20.02 & 20.99\% & 1.51m & 31.15 & 34.75\% & 3.18m & 141.68 & 97.39\% & 5.99m \\
    \multicolumn{1}{l}{AM} & RL+BS & 19.53 & 18.03\% & 21.99m & 29.90 & 29.23\% & 1.64h & 129.4 & 80.28\% & 1.81h \\
    \multicolumn{1}{l}{GCN} & SL+G  & 29.72 & 79.61\% & 6.67m & 48.62 & 110.29\% & 28.52m & N/A   & N/A   & N/A \\
    \multicolumn{1}{l}{GCN} & SL+BS & 30.37 & 83.55\% & 38.02m & 51.26 & 121.73\% & 51.67m & N/A   & N/A   & N/A \\
    \multicolumn{1}{l}{POMO+EAS-Emb(ICLR2022)} & RL+AS & 19.24 & 16.25\% & 12.80h & N/A   & N/A   & N/A   & N/A   & N/A   & N/A \\
    \multicolumn{1}{l}{POMO+EAS-Lay} & RL+AS & 19.35 & 16.92\% & 16.19h & N/A   & N/A   & N/A   & N/A   & N/A   & N/A \\
    \multicolumn{1}{l}{POMO+EAS-Tab} & RL+AS & 24.54 & 48.22\% & 11.61h & 49.56 & 114.36\% & 63.45h & N/A   & N/A   & N/A \\
    \multicolumn{1}{l}{Att-GCN(AAAI2021)} & SL+MCTS & 16.97 & 2.54\% & 2.20m & 23.86 & 3.22\% & 4.10m & 74.93 & 4.39\% & 21.49m \\
    \multicolumn{1}{l}{DIMES(NeurIPS2022)} & RL+S  & 18.84 & 13.84\% & 1.06m & 26.36 & 14.01\% & 2.38m & 85.75 & 19.48\% & 4.80m \\
    \multicolumn{1}{l}{DIMES} & RL+AS+S & 17.8  & 7.55\% & 2.11h & 24.89 & 7.70\% & 4.53h & 80.42 & 12.05\% & 3.12h \\
    \multicolumn{1}{l}{DIMES} & RL+MCTS & 16.87 & 1.93\% & 2.92m & 23.73 & 2.64\% & 6.87m & 74.63 & 3.98\% & 29.83m \\
    \multicolumn{1}{l}{DIMES} & RL+AS+MCTS & 16.84 & 1.76\% & 2.15h & 23.69 & 2.46\% & 4.62h & 74.06 & 3.19\% & 3.57h \\
    \multicolumn{1}{l}{H-TSP(AAAI2023)} & RL & 17.59 & 6.28\% & 27.35s & 24.66 & 6.66\% & 56.12s & 77.74 & 8.31\% & 57.48s \\
    \midrule
    \multirow{4}[1]{*}{DTTGF+Att-GCN(ours)} & SL+S+2-OPT & 17.19  & 3.84\% & 1.12m & 24.01  & 3.85\% &  2.90m & 76.24 & 6.23\% &  11.29m \\
          & SL+WU+S+2-OPT & 16.96  & 2.47\% & 1.12m+2.70m & 23.83  & 3.05\% &  2.90m+15.44m & 75.09 & 4.63\% &  11.29m+1.58h \\
          & SL+MCTS & 16.74  & \textbf{1.13\%} &  2.30m & 23.45  & \textbf{1.43\%} & 4.15m & 73.57 & \textbf{2.51\%} & 24.74m \\
          & SL+WU+MCTS & 16.71  & \textbf{0.99\%} &  2.30m+2.70m & 23.44  & \textbf{1.39\%} & 4.15m+15.44m & 73.48 & \textbf{2.38\%} & 24.74m+1.58h \\
    \midrule
    \multirow{4}[1]{*}{DTTGF+POMO(ours)} & RL+S+2-OPT & 17.17  & 3.74\% &  2.86m & 24.08  & 4.14\% &  5.37m & 78.01 & 8.69\% &  13.55m \\
          & RL+WU+S+2-OPT & 1.03  & 2.50\% &  2.86m+2.70m & 23.87  & 3.23\% &  5.37m+15.44m & 75.87 & 5.71\% &  13.55m+1.58h \\
          & RL+MCTS & 24.77  & 9.40\% &  4.14m & 26.82  & 16.01\% &  6.34m & 107.44 & 49.70\% &  23.90m \\
          & RL+WU+MCTS & 16.82  & \textbf{1.66\%} &  4.14m+2.70m & 23.70  & 2.50\% &  6.34m+15.44m & 73.56 & \textbf{2.49\%} &  23.90m+1.58h \\
    \bottomrule
    \end{tabular}%
            \begin{tablenotes}    
        \footnotesize               
        \item[1] Where the symbol $*$ represents the benchmark algorithm, while "length" indicates the average solution length across all instances.          
        \item[2] "Time" reflects the cumulative computational duration for all instances, and "drop" signifies the deviation from the benchmark.        
        \item[3]  The notation "WU" designates the warm-up module, with the supplementary time incorporated in methods employing the warm-up module representing the duration required for warm-up procedures.        
        \item[4] The decoding scheme in each method (if applicable) is further specified as Greedy Decoding (G), Sampling (S), Beam Search (BS) and Monte Carlo Tree Search (MCTS).        
      \end{tablenotes}            
    \end{threeparttable}       
    }
  \label{TSP_re}%
\end{table*}%

\begin{table*}[!h]
  \centering
  \caption{Results of DTTGF(embedding AM and GCM) w.r.t. AM and GCN, tested on TSP-500, TSP-1000 and TSP-10000.}
  \scalebox{0.65}{
    \begin{tabular}{ll|ccc|ccc|ccc}
    \toprule
    \multicolumn{1}{c}{\multirow{2}[2]{*}{Method}} & \multicolumn{1}{c|}{\multirow{2}[2]{*}{Type}} & \multicolumn{3}{c|}{TSP-500} & \multicolumn{3}{c|}{TSP-1000} & \multicolumn{3}{c}{TSP-10000} \\
          &       & \multicolumn{2}{c}{Drop↓} & Time↓ & \multicolumn{2}{c}{Drop↓} & Time↓ & \multicolumn{2}{c}{Drop↓} & Time↓ \\
    \midrule
    AM    & RL+S  & \multicolumn{2}{c}{36.84\%} & 15.64m & \multicolumn{2}{c}{85.15\%} & 63.97m & \multicolumn{2}{c}{501.27\%} & 12.63m \\
    AM    & RL+G  & \multicolumn{2}{c}{20.99\%} & 1.51m & \multicolumn{2}{c}{34.75\%} & 3.18m & \multicolumn{2}{c}{97.39\%} & 5.99m \\
    AM    & RL+BS & \multicolumn{2}{c}{18.03\%} & 21.99m & \multicolumn{2}{c}{29.23\%} & 1.64h & \multicolumn{2}{c}{80.28\%} & 1.81h \\
    \midrule
    \multirow{4}[1]{*}{DTTGF+AM(ours)} & RL+S+2-OPT & \multicolumn{2}{c}{\textbf{5.60\%}} &  4.75m & \multicolumn{2}{c}{\textbf{5.62\%}} &  10.63m & \multicolumn{2}{c}{\textbf{12.40\%}} &  1.03h \\
          & RL+WU+S+2-OPT & \multicolumn{2}{c}{\textbf{3.98\%}} &  4.75m+2.70m & \multicolumn{2}{c}{\textbf{4.59\%}} &   10.63m+15.44m & \multicolumn{2}{c}{\textbf{11.04\%}} &  1.03h+1.58h \\
          & RL+MCTS & \multicolumn{2}{c}{\textbf{12.69\%}} &  6.11m & \multicolumn{2}{c}{\textbf{20.74\%}} &  11.83m & \multicolumn{2}{c}{\textbf{54.26\%}} &  51.05m \\
          & RL+WU+MCTS & \multicolumn{2}{c}{\textbf{2.68\%}} &  6.11m+2.70m & \multicolumn{2}{c}{\textbf{4.28\%}} &  11.83m+15.44m & \multicolumn{2}{c}{\textbf{27.55\%}} & 51.05m+1.58h \\
    \midrule
    GCN   & SL+G  & \multicolumn{2}{c}{79.61\%} & 6.67m & \multicolumn{2}{c}{110.29\%} & 28.52m & \multicolumn{2}{c}{N/A} & N/A \\
    GCN   & SL+BS & \multicolumn{2}{c}{83.55\%} & 38.02m & \multicolumn{2}{c}{121.73\%} & 51.67m & \multicolumn{2}{c}{N/A} & N/A \\
    \midrule
    \multirow{4}[1]{*}{DTTGF+GCN(ours)} & SL+S+2-OPT & \multicolumn{2}{c}{\textbf{6.44\%}} & 1.42m & \multicolumn{2}{c}{\textbf{6.45\%}} & 3.16m & \multicolumn{2}{c}{\textbf{14.93\%}} & 1.15h \\
          & SL+WU+S+2-OPT & \multicolumn{2}{c}{\textbf{5.01\%}} & 1.42m+2.70m & \multicolumn{2}{c}{\textbf{5.57\%}} & 3.16m+15.44m & \multicolumn{2}{c}{\textbf{11.05\%}} & 1.15h+1.58h \\
          & SL+MCTS & \multicolumn{2}{c}{\textbf{1.50\%}} & 3.38m & \multicolumn{2}{c}{\textbf{2.13\%}} & 5.37m & \multicolumn{2}{c}{\textbf{3.64\%}} & 1.00h \\
          & SL+WU+MCTS & \multicolumn{2}{c}{\textbf{2.10\%}} & 3.38m+2.70m & \multicolumn{2}{c}{\textbf{2.64\%}} & 5.37m+15.44m & \multicolumn{2}{c}{\textbf{3.84\%}} & 1.00h+1.58h \\
    \bottomrule
    \end{tabular}%
    }
  \label{Freamwork}%
\end{table*}%

\subsection{Baselines}
The baselines employed in the experiments are of two types: the traditional methods and the learning-base methods.


Traditional methods: we have Concorde \cite{Applegate_Bixby_Chvátal_Cook_2007}, a classic TSP accurate solver, and Gurobi \footnote{See https://www.gurobi.com}, a widely used industrial combinatorial optimization problem solver. Also included is LKH-3 \cite{36b9628e7a874d208624584d8a470985}, the classic heuristic TSP solver, and Farthest Insertion, a heuristic solver for most combinatorial optimization problems.

Learning-based methods:EAN\cite{DBLP:conf/cpaior/DeudonCLAR18} (CPAIOR2018) is a re-engineered LSTM neural network framework that introduced 2-OPT assisted search. AM\cite{kool2018attention} (ICLR2019) brought Attention Mechanisms to TSP solving. GCN\cite{joshi2019efficient} focuses on constructing and solving TSP representations with graph neural networks. POMO+EAS\cite{DBLP:conf/iclr/HottungKT22} (ICLR2022) improved training speed and extended the existing TSP solver with a new combined search strategy. Att-GCN (AAAI2021) solves TSPs using supervised learning and reinforcement learning. DIMES (NeurIPS2022) provides a new spatial representation and meta-learning framework for solving TSP instances of different sizes. Lastly, H-TSP (AAAI2023)\cite{DBLP:conf/aaai/PanJDF0S023} jointly trains upper and lower layer solutions that can directly generate solutions for a given TSP instance without relying on any time-consuming search process.


\subsection{Comparative Study}
In our framework-based study, Att-GCN and POMO were integrated as the experimental TSP solvers. Att-GCN requires a pre-trained GCN model for heatmap generation, representing a two-stage learning approach. POMO excels in smaller TSP instances and is included as a one-stage solver in our framework.

Table \ref{TSP_re} shows DTTGF's advantages over other learning-based methods in scheduling and timing, with all baseline results from \cite{qiu2022dimes} except Att-GCN's. Our framework is easily integrated by embedding a TSP solver without extra training. Using MCTS, DTTGF outperforms Att-GCN and DIMES. With S+2-OPT, a fast search method for real-time tasks, DTTGF's accuracy drops but time is reduced. DTTGF+S+2-OPT even surpasses MCTS-based methods in accuracy, including for 1000-point datasets.

A key issue is POMO's scaling challenges with Large-Scale TSPs (LSTSPs). The POMO+EAS benchmark struggled with TSP-1000 and TSP-10000 within 8 hours. However, integrated with DTTGF, POMO outperforms specialized LSTSP solvers like DIMES across all datasets. Details on the framework's generalizability and reliability are in the ablation study.

\subsection{Ablation Study}
Ablation results in Table I show the proposed assumptions' validity against DT. Att-GCN+DTTGF (without WU) differs from Att-GCN in using DT-based sampling and fusion. DTTGF surpasses Att-GCN in accuracy with MCTS decoding, suggesting DT's enhancement potential.

DTTGF vs. DTTGF(warm-up) shows warm-up boosts accuracy, especially for one-stage methods. With MCTS, DTTGF with POMO improves TSP-1000 by over 10\%  .

Warm-up times for TSP-500, TSP-1000, and TSP-10000 are 1.22s, 7.23s, and 5.12 min, respectively.

AM and GCN integrated into DTTGF (Table II) show significant LSTSP improvements in accuracy and efficiency. GCN, like POMO, reaches TSP-10000, a feat beyond the original approach, proving DTTGF's versatility in enhancing existing methods.

\section{Conclusion}
Introducing DTTGF, a novel framework upgrading TSP solvers for UAVRP with many sites. It uses Delaunay triangulation for graph decomposition and embeds current solvers. With a warm-up strategy, DTTGF extends solvers efficiently, enhancing generalizability and scalability. Embedding Att-GCN and POMO, it handles LSTSP, outperforming or matching SOTA on three datasets. Ablation studies show DT-based subgraph methods preserve features and boost accuracy. The warm-up strategy improves search accuracy within reasonable time. Future work could enhance the framework's effectiveness, especially for large TSP instances, and refine the warm-up strategy with adaptive hyperparameters. The framework may also apply to other combinatorial optimization problems.




\bibliographystyle{ieeetr}
\bibliography{reference}

\end{document}